%% file: 1-main.tex
\documentclass{article} 
\usepackage{iclr2027_conference,times}

\input{math_commands.tex}

\usepackage{hyperref}
\usepackage{url}

\usepackage{microtype}
\usepackage{graphicx}
\usepackage{subcaption}
\usepackage{booktabs} 

\usepackage{amsmath}
\usepackage{amssymb}
\usepackage{mathtools}
\usepackage{amsthm}
\usepackage[table]{xcolor}
\usepackage{multirow}
\usepackage{makecell}
\usepackage{wrapfig}

\title{Reciprocal Guidance: Orchestrating Draft and Verify Budgets for Advancing the Diffusion-AR Self-Speculation Frontier}

\author{
Linye Wei\textsuperscript{\rm 1,2,*}\quad
Shutian Zheng\textsuperscript{\rm 3,*}\quad
Haoyu Zeng\textsuperscript{\rm 4}\quad
Meng Li\textsuperscript{\rm 1,2,5,\textdagger}
\\
\textsuperscript{\rm 1}Institute for Artificial Intelligence, Peking University\\
\textsuperscript{\rm 2}School of Integrated Circuits, Peking University\\
\textsuperscript{\rm 3}College of Engineering, Peking University\\
\textsuperscript{\rm 4}School of Airspace Science and Engineering, Shandong University\\
\textsuperscript{\rm 5}Beijing Advanced Innovation Center for Integrated Circuits
}

\iclrfinalcopy 
\begin{document}

\maketitle

\begingroup
\renewcommand{\thefootnote}{}
\footnotetext{%
  \textsuperscript{*}Equal contribution.
  \textsuperscript{\textdagger}Corresponding author: meng.li@pku.edu.cn.
}
\endgroup

\begin{abstract}
Diffusion drafting with autoregressive (AR) verification has emerged as a promising paradigm for efficient speculative decoding. Recent self-speculation models, represented by Nemotron-Labs-Diffusion, further simplify the speculative pipeline by unifying drafting and verification within a shared backbone, while enabling longer acceptance lengths. However, the Pareto frontier between aggregate and per-request throughput remains underexplored. At low concurrency, sequential draft-verify execution requires two model forward passes per round, limiting the effective tokens per forward (TPF). By contrast, at high concurrency, longer drafts incur increasingly expensive computation, forcing individual requests to operate under constrained speculation budgets and preventing full exploitation of the full-backbone drafter. Our key observation indicates that drafting and verification exhibit reciprocal predictability. Draft logits can anticipate likely verification mismatches, while recent verification outcomes predict future drafting utility and suitable block sizes. Building on this observation, we introduce~\textbf{Reciprocal Guidance} (\textbf{RecGuide}), a runtime draft-verify orchestration framework that adapts speculative decoding to varying serving loads. RecGuide exploits spare compute capacity through verification-overlapped drafting at low concurrency, while dynamically allocating request-specific draft block sizes as the workload becomes increasingly compute-intensive. Experiments across a wide range of concurrency levels demonstrate consistent throughput improvements over vanilla self-speculation, achieving up to 1.8× speedup.
\end{abstract}

\input{docs/1-introduction}
\input{docs/2-related-work}
\input{docs/3-preliminary-and-bottleneck}
\input{docs/4-reciprocal-predictability}

\input{docs/5-recguide}
\input{docs/6-experiments}

\section{Conclusion}
This work presents Reciprocal Guidance (RecGuide), a runtime draft-verify orchestration framework for diffusion-AR self-speculation. RecGuide leverages reciprocal predictability between drafting and verification to overlap the two stages at low concurrency and adapt request-specific draft budgets under heavier workloads. Across a wide range of concurrency levels, RecGuide consistently improves throughput over vanilla LinearSS, achieving up to 1.8× speedup. These results highlight self-speculation that reuses the full target backbone as the drafter as a promising alternative for efficient speculative decoding across diverse serving loads and concurrency regimes.

\bibliography{2-reference}
\bibliographystyle{iclr2027_conference}


\end{document}

%% file: math_commands.tex
\usepackage{amsmath,amsfonts,bm}

\def\eqref#1{equation~\ref{#1}}

\def\1{\bm{1}}

\DeclareMathAlphabet{\mathsfit}{\encodingdefault}{\sfdefault}{m}{sl}
\SetMathAlphabet{\mathsfit}{bold}{\encodingdefault}{\sfdefault}{bx}{n}



%% file: docs/1-introduction.tex
\section{Introduction}
\label{sec:introduction}

Speculative decoding \citep{chen2023acceleratinglargelanguagemodel,leviathan2023fast} has emerged as an important technique for accelerating large language model (LLM) inference. It employs a lightweight draft model to autoregressively generate multiple candidate tokens, which are then verified and potentially accepted in a single forward pass. Recently, inspired by diffusion language models (dLLMs) \citep{nie2026large,ye2025dream7bdiffusionlarge,cheng2026sdar}, diffusion-based speculative decoding \citep{chen2026dflash,huang2026domino,cheng2026dspark} uses a predictor with bidirectional attention to generate draft tokens in parallel, followed by autoregressive verification. By reducing sequential drafting overhead while maintaining high acceptance rates, this paradigm has attracted growing attention.

Alongside this line of work, self-speculation methods such as Nemotron-Labs-Diffusion \citep{fu2026nemotronlabsdiffusion,liu2026tidar} use a single shared backbone and switch between bidirectional and causal attention to support diffusion drafting and autoregressive verification without requiring a separate drafter. This design simplifies the speculative pipeline \citep{zhang2026flexdraftflexiblespeculativedecoding} and, by reusing the full target backbone itself as the drafter \citep{zhang2024draft}, offers stronger task generalization and the potential for longer acceptance lengths than lightweight draft models.

However, concurrency varies dynamically in real-world serving workloads \citep{chen2024sequoia}, while current self-speculation typically relies on fixed draft lengths (referred to as block sizes hereafter) tailored to different concurrency regimes. We argue that this coarse-grained allocation is insufficient \citep{sadhukhan2025magicdec}. \textbf{At low concurrency}, the sequential draft-then-verify pipeline requires two expensive forward passes per speculative round, reducing the effective tokens per forward (TPF) to roughly half of the acceptance length and substantially diminishing the gains enabled by the full-backbone drafter. \textbf{At high concurrency}, by contrast, maintaining a large block size for every request becomes impractical, as the resulting computation pushes GPU execution into the compute-bound regime and significantly increases forward-pass latency. Yet choosing a small block size underutilizes self-speculation's potential for longer acceptance lengths compared with conventional speculative decoding paradigms. This motivates the following question:

\textit{Can we orchestrate draft and verify budgets to advance the aggregate and per-request throughput frontier, thereby making self-speculation a competitive paradigm for speculative decoding?}

To address this challenge, we propose~\textbf{Reciprocal Guidance} (\textbf{RecGuide}), grounded in our core observation that drafting and verification on the shared backbone are not independent, but exhibit strong reciprocal predictability. On the one hand, draft logits provide informative signals for anticipating likely mismatch positions in subsequent AR verification, as well as a likely correction token. On the other hand, recent verification outcomes reflect the local drafting difficulty of each request and provide an effective signal for selecting an appropriate block size in the next round.

\begin{figure}[t]
    \centering
    \includegraphics[width=0.98\linewidth]{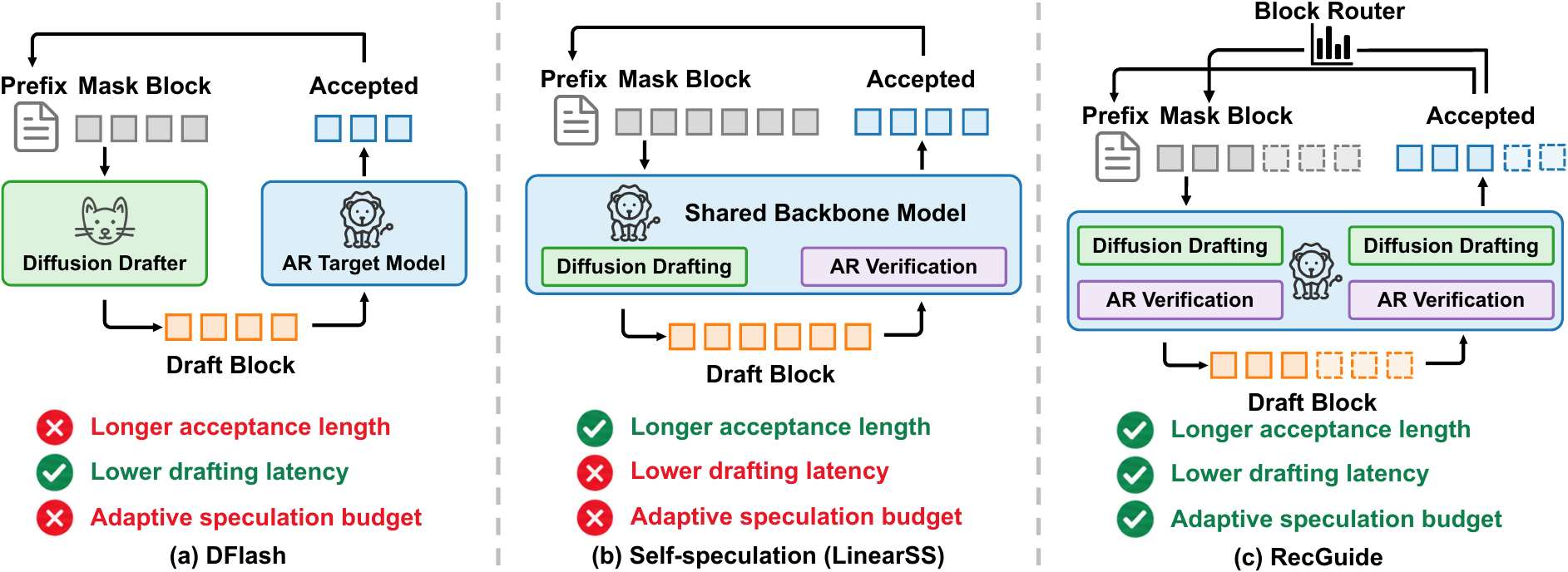}
    \caption{Comparison between RecGuide and existing diffusion-based speculative decoding.}
    \label{fig:overview}
\end{figure}

Building on these observations, RecGuide provides a draft-verify orchestration framework for self-speculation with complementary strategies across varying concurrency levels, as illustrated in Figure~\ref{fig:overview}. At low concurrency, RecGuide predicts the upcoming verification outcome from the draft probability distribution and co-batches likely next-round drafts with AR verification. This overlap is particularly natural for self-speculation because drafting and verification share the same weights, enabling simple co-batching without model switching or separate devices while exploiting spare compute capacity in the memory-bound regime \citep{liu2025pearl,kumar2026speculativespeculativedecoding}. As the workload increases, RecGuide instead uses verification feedback to estimate future drafting utility and dynamically assigns request-specific draft block sizes. Together with bucketed serving based on block size, this strategy balances the drafting capability of the model against available GPU compute.

We evaluate RecGuide under diverse serving regimes. Our method consistently improves throughput over vanilla self-speculation, enhancing speculative efficiency under light workloads and serving efficiency under heavy workloads. Our main contributions can be summarized as follows:
\begin{itemize}
  \item We characterize the efficiency bottlenecks and pipeline challenges of self-speculation, and uncover reciprocal predictability between drafting and verification.
  \item We propose RecGuide, which orchestrates draft and verify budgets across varying serving loads through predictive draft overlap and adaptive block sizing.
  \item Extensive experiments demonstrate that RecGuide achieves consistent throughput improvements over vanilla self-speculation, with up to a 1.8× speedup.
\end{itemize}

%% file: docs/2-related-work.tex
\section{Related work}
\label{sec:related work}

\textbf{Drafting Paradigms for Speculative Decoding.}
Speculative decoding \citep{chen2023acceleratinglargelanguagemodel,leviathan2023fast,pmlr-v235-cai24b,zhong2024propd} has become a widely adopted approach for accelerating large language model inference. It uses a lightweight drafter to propose multiple future tokens, which are verified in parallel by the target model, enabling multiple tokens to be accepted per target-model forward without altering the generation distribution. Early methods relied on smaller standalone models sharing the target vocabulary, making speedup sensitive to drafter latency. EAGLE series \citep{pmlr-v235-li24bt,li2024eagle,li2026eagle} improve drafting efficiency by exploiting target model representations. Nevertheless, autoregressive drafting still incurs latency that grows with draft length. Inspired by diffusion \citep{ho2020denoising,yan2026less,yan2025entropy} and diffusion language models \citep{nie2026large,ye2025dream7bdiffusionlarge,cheng2026sdar,bie2025llada20scalingdiffusionlanguage,wei2026orchestratingdualboundariesarithmeticintensity}, recent approaches \citep{chen2026dflash,huang2026domino,cheng2026dspark} employ diffusion drafters with bidirectional attention to generate a draft block in a single forward pass, reducing sequential drafting overhead while achieving higher acceptance rates. Meanwhile, self-speculative methods \citep{fu2026nemotronlabsdiffusion,liu2026tidar,zhang2026flexdraftflexiblespeculativedecoding} further unify diffusion drafting and autoregressive verification within a shared model, enabling stronger drafting capability and longer acceptance lengths at increased drafting cost. Our work focuses on the runtime execution and efficiency of this paradigm.

\textbf{Runtime Orchestration for Speculative Decoding.}
Beyond drafter design, recent work has increasingly optimized speculative decoding through parallel execution and adaptive runtime strategies \citep{zhang2026dflowenablingverifierinformation,lin2026retracerejectedtrajectoryconditioningspeculative}, aiming to break the sequential draft-then-verify dependency and accommodate diverse serving conditions. PEARL \citep{liu2025pearl} overlaps drafting and verification while adapting draft length to mitigate mutual waiting between a separate drafter and verifier. SSD \citep{kumar2026speculativespeculativedecoding} further predicts likely verification outcomes before verification completes and asynchronously prepares future speculation. In contrast, our proposed RecGuide targets self-speculation where both stages invoke the same backbone, allowing future drafts to be directly co-batched with ongoing verification without model switching or additional devices. This turns the efficiency bottleneck into a limited-compute allocation problem, especially at high concurrency. At the drafting stage, prior work \citep{sadhukhan2025magicdec,ning2026cas,zhang2026learning} explores more flexible strategies to dynamically determine the amount and configuration of speculation for higher throughput. At the verification stage, another line of work \citep{liu2026dcutadaptiveverificationdepth,cheng2026dspark,liu2026angelspecrealworldhighperformance} selectively allocates verification effort based on confidence signals or hardware-aware runtime models. These two stages, however, are typically optimized separately. RecGuide instead treats drafting and verification as mutually informative and coupled stages, using bidirectional signals to jointly orchestrate their budgets across concurrency regimes.

%% file: docs/3-preliminary-and-bottleneck.tex
\section{Preliminary and Bottleneck}
\label{sec:preliminary and bottleneck}

\subsection{Self-Speculation with a Shared Backbone}
\label{sec:preliminary}

Unlike conventional speculative decoding, which employs a lightweight drafter alongside the target model, shared-backbone self-speculation jointly optimizes autoregressive and diffusion objectives so that a single backbone supports both diffusion drafting and autoregressive verification, as illustrated in Figure~\ref{fig:overview}(b). Given a generated prefix, the model first predicts multiple future tokens in parallel under the diffusion mode and then switches to the autoregressive mode for causal verification following standard speculative decoding, accepting the correct prefix up to the first mismatch.

The two modes can be switched simply by changing the attention pattern. Let the current prefix length be $P$ and the draft block size be $B$. We denote prefix positions by $\mathcal{P}={1,\ldots,P}$ and draft positions by $\mathcal{D}={P+1,\ldots,P+B}$. Under diffusion drafting, the prefix preserves causal dependency, while tokens in $\mathcal{D}$ can attend to the prefix and entire draft region, enabling parallel prediction through bidirectional attention. Autoregressive verification instead enforces strict causal attention, where position $i$ can only attend to positions $j\le i$. The corresponding attention masks are as follows:

\begin{equation}
M_{ij}^{\mathrm{Diff}}
=
\begin{cases}
0, & i\in\mathcal{P},\ j\in\mathcal{P},\ j\le i,\\
0, & i\in\mathcal{D},\ j\in\mathcal{P}\cup\mathcal{D},\\
-\infty, & \text{otherwise},
\end{cases}
\qquad
M_{ij}^{\mathrm{AR}}
=
\begin{cases}
0, & j\le i,\\
-\infty, & j>i.
\end{cases}
\label{eq:attention_masks}
\end{equation}

Consequently, $M_{ij}^{\mathrm{Diff}}$ enables bidirectional interaction within the draft region, whereas $M_{ij}^{\mathrm{AR}}$ restores left-to-right dependency. This switching allows the full backbone itself to serve as the drafter without deploying a separate draft model. 
However, sharing the backbone also directly couples drafting and verification at runtime, forcing the two full-backbone stages to execute sequentially within each speculative round and thereby creating efficiency characteristics distinct from conventional speculative decoding.
We next examine the resulting bottlen  ecks under different serving conditions.

\subsection{Bottlenecks across Serving Loads}
\label{sec:bottleneck}

\begin{figure}[t]
    \centering

    \begin{subfigure}[t]{0.45\linewidth}
        \centering
        \includegraphics[
            width=\linewidth
        ]{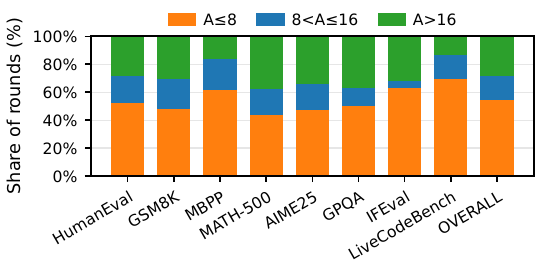}
        \caption{}
        \label{fig:bottleneck-a}
    \end{subfigure}
    \hfill
    \begin{subfigure}[t]{0.45\linewidth}
        \centering
        \includegraphics[
            width=\linewidth
        ]{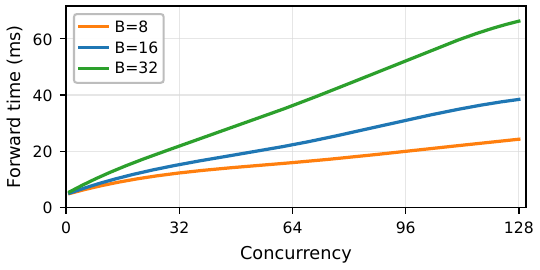}
        \caption{}
        \label{fig:bottleneck-b}
    \end{subfigure}

    \caption{Acceptance variability and block-dependent forward cost of self-speculation. (a) Acceptance length distribution with $B=32$. (b) Forward time across concurrency levels for $B\in\{8,16,32\}$.}
    \label{fig:bottleneck}
\end{figure}

To characterize the efficiency of self-speculation across serving loads, we profile Nemotron-Labs-Diffusion 8B in terms of acceptance behavior and forward cost. As shown in Figure~\ref{fig:bottleneck}(a), with block size $32$, we report the fraction of rounds whose acceptance length $A$ falls into $A\leq8$, $8<A\leq16$, and $A>16$ across benchmarks. Nearly $30\%$ of the rounds accept more than $16$ tokens, revealing strong potential for long drafts, while over half fail within the first $8$ positions, indicating substantial variation in useful drafting length. We further examine forward time under fixed block sizes of $8$, $16$, and $32$ as concurrency increases on a single B200 GPU, as illustrated in Figure~\ref{fig:bottleneck}(b). Forward time grows with concurrency, with larger block sizes incurring increasingly higher latency under heavier workloads. Together, these profiles reveal a mismatch between highly variable acceptance potential and load-dependent drafting cost across concurrency regimes.

\textbf{Low-concurrency Inefficiency.}
At low concurrency, inefficiency comes from both pipeline serialization and wasted drafting. Drafting and verification execute sequentially, causing mutual waiting that is often negligible with lightweight drafters but costly when both stages invoke the full backbone. Moreover, Figure~\ref{fig:bottleneck}(a) shows that large blocks can waste substantial drafting computation when mismatches occur early. Fortunately, execution remains largely memory-bound at low concurrency, leaving spare compute that can be used to overlap drafting with verification and proactively correct likely mismatches.

\textbf{High-concurrency Inefficiency.}
As concurrency increases, Figure~\ref{fig:bottleneck}(b) shows that block size becomes increasingly constrained by compute load, favoring smaller blocks. However, uniformly using a small block remains suboptimal, since it sacrifices the long-acceptance potential observed in Figure~\ref{fig:bottleneck}(a). Since GPU latency typically grows sublinearly with the number of processed tokens, additional drafting remains beneficial when the acceptance gain outweighs the corresponding latency overhead. This motivates dynamically selecting the block size for each request rather than using a fixed global budget. The resulting divergence in per-request computation lengths will be handled through bucketed serving, as discussed in Section~\ref{sec:implementation}.

%% file: docs/4-reciprocal-predictability.tex
\section{Reciprocal Predictability between Drafting and Verification}
\label{sec:reciprocal predictability}

The bottlenecks above show that runtime orchestration must account for both varying serving loads and the dynamic nature of decoding. Shared-backbone execution exposes complementary signals from drafting and verification that can guide such adaptation. We next establish this~\textbf{reciprocal predictability} and its implications for runtime decisions.

\subsection{Draft Distributions Predict Verification Failures}
\label{sec:motivation-1}

\begin{wrapfigure}{r}{0.46\linewidth}
    \centering
    \vspace{-0.8\baselineskip}
    \includegraphics[width=\linewidth]{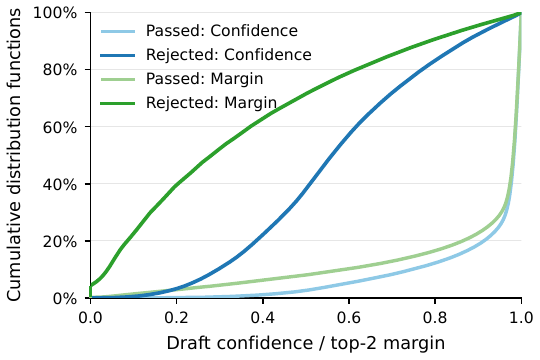}
    \caption{Cumulative distribution functions (CDFs) of draft confidence and top-2 probability margin during drafting rounds.}
    \label{fig:draft_to_verify_predictability}
\end{wrapfigure}

We first examine whether the draft distribution contains predictive cues about subsequent verification. Under greedy decoding, we define the confidence of the drafted token at position $i$ as $C_i=P_{\mathrm{1st}}(i)$. Figure~\ref{fig:draft_to_verify_predictability} compares the cumulative distribution functions (CDFs) of $C_i$ at verifier-passed and first-rejected positions across math, code, and QA benchmarks. Passed tokens are heavily concentrated at high confidence values, whereas first-rejected positions exhibit a substantially broader distribution shifted toward lower confidence. This clear separation reveals that draft confidence can anticipate likely verification failures before the verifier is executed.

We further consider the probability margin between the two most likely draft candidates, $M_i=P_{\mathrm{1st}}(i)-P_{\mathrm{2nd}}(i)$. Its CDF exhibits a similar separation between passed and first-rejected positions. A small margin not only reflects uncertainty in the submitted draft token, but also shows that the second-ranked candidate remains competitive. Although this candidate is not necessarily correct, it offers a natural alternative at likely mismatch positions. 
These observations imply that likely verification failures can be anticipated and prepared for before causal verification completes.

\subsection{Verification History Predicts Future Drafting Utility}
\label{sec:motivation-2}

\begin{wraptable}{r}{0.46\linewidth}
    \centering
    \vspace{-0.8\baselineskip}
    \caption{Transition probabilities of acceptance intervals between adjacent decoding rounds.}
    \label{tab:verify_to_draft_predictability}

    \small
    \setlength{\tabcolsep}{5.0pt}
    \renewcommand{\arraystretch}{1.08}

    \begin{tabular}{c c c c}
        \toprule
        $A_t \backslash A_{t+1}$
        & $[1,8]$
        & $(8,16]$
        & $(16,32]$ \\
        \midrule
        $[1,8]$
        & \textbf{70.30\%}
        & 18.11\%
        & 11.60\% \\
        $(8,16]$
        & 58.39\%
        & \textbf{22.42\%}
        & 19.19\% \\
        $(16,32]$
        & 20.83\%
        & 12.51\%
        & \textbf{66.66\%} \\
        \bottomrule
    \end{tabular}

\end{wraptable}

Conversely, we examine whether completed verification is predictive of how far the next round should draft. Without introducing a specific utility metric, we analyze the temporal correlation of acceptance lengths across adjacent rounds, as shown in Table~\ref{tab:verify_to_draft_predictability}. To expose the model's full acceptance potential, the block size is fixed at $32$, and the acceptance length at round $t$ is denoted by $A_t$. We divide $A_t$ into three intervals: short acceptance with $A_t\leq8$, medium acceptance with $8<A_t\leq16$, and long acceptance with $A_t>16$. These intervals align with the candidate block sizes $8$, $16$, and $32$. Table~\ref{tab:verify_to_draft_predictability} reports the conditional distribution of $A_{t+1}$ over the three intervals. Both short and long acceptance exhibit clear temporal persistence, as the next round is most likely to remain in the same interval. This pattern suggests that local generation difficulty tends to persist across adjacent rounds, making verification history a useful prior for future drafting budgets.

This persistence, however, is not static. Table~\ref{tab:verify_to_draft_predictability} also shows frequent transitions between adjacent rounds. In particular, only $22.42\%$ of medium acceptance rounds remain in the same interval, primarily because short acceptance rounds dominate all decoding rounds. Thus, each response exhibits both local persistence and regime transitions. A fixed block size therefore cannot consistently match the local drafting utility, calling for adaptive block size allocation under limited compute.

%% file: docs/5-recguide.tex
\section{RecGuide: Runtime Draft-Verify Orchestration}
\label{sec:recguide}

\begin{figure}[t]
    \centering
    \includegraphics[width=0.98\linewidth]{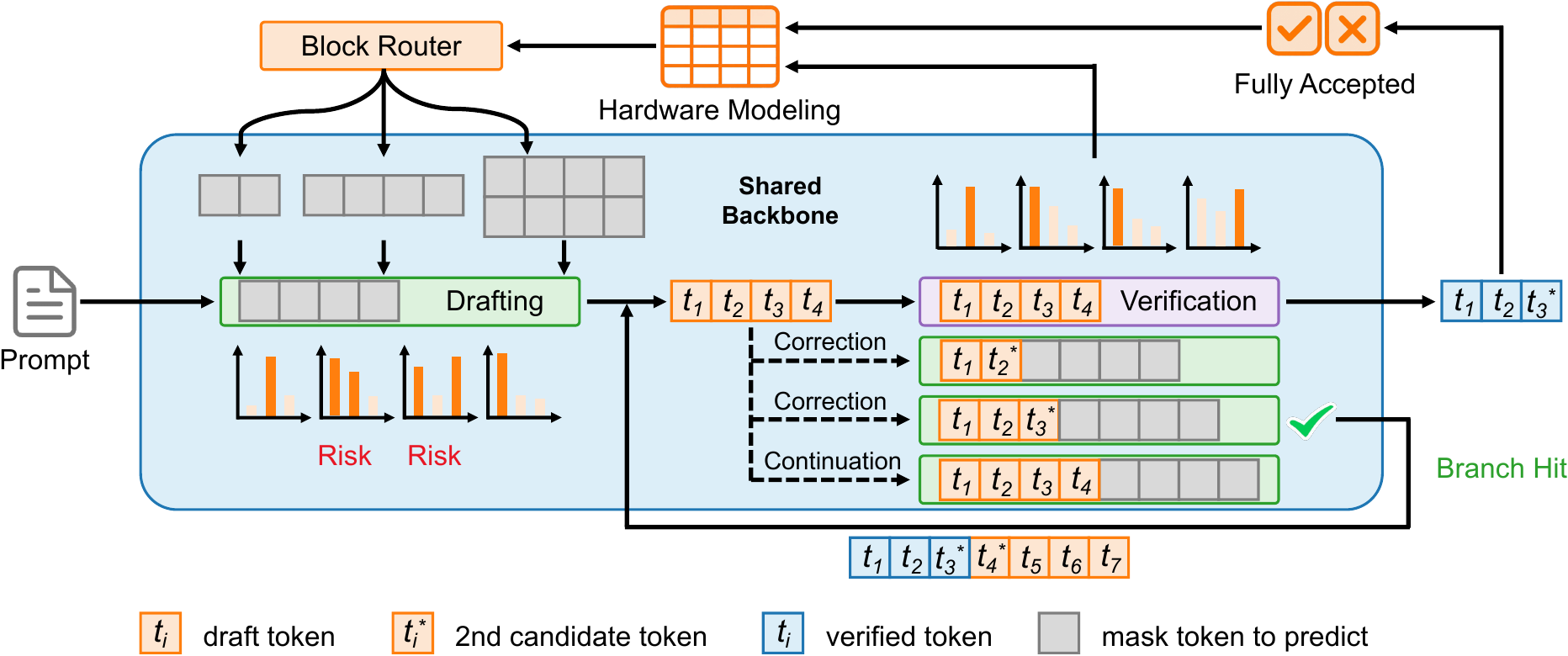}
    \caption{Overview of RecGuide runtime orchestration. Draft uncertainty guides predictive draft-verify overlap through correction and continuation branches, while verification feedback and hardware profiles guide request-specific block size allocation.}
    \label{fig:recguide}
\end{figure}

RecGuide converts reciprocal predictability into runtime decisions, as illustrated in Figure~\ref{fig:recguide}. Under light workloads, spare compute prepares likely future drafts alongside verification. Once compute becomes limiting, verification history instead guides each request's drafting budget.

\subsection{Predictive Draft-Verify Overlap}
\label{sec:method-1}

As discussed before, the probability margin $M_i=P_{\mathrm{1st}}(i)-P_{\mathrm{2nd}}(i)$ reflects draft uncertainty and second-ranked token competitiveness. We mark positions with $M_i<\delta$ as potential mismatches and forms a correction branch by replacing the draft token with its second candidate.
Because later positions in a parallel draft are generally less reliable and less valuable for correcting the first mismatch, RecGuide retains only the earliest two risky positions. When at most two such positions exist, it additionally prepares a continuation branch assuming full acceptance of the current block. Together with the verifier, each round contains at most four logical rows, bounding extra computation. 

These branches prepare drafts for the next round in parallel with current verification. A correction branch is reused only if the verifier rejects at the predicted position and produces the same replacement token, whereas retaining the continuation branch requires full acceptance of the current block and a match between its first token and the verifier bonus token. Unmatched branches are discarded. A confirmed draft can immediately repeat the procedure in the next round, enabling recursive overlap. 

\subsection{Verification Guided Adaptive Drafting}
\label{sec:method-2}

At higher concurrency, costly prospective branches shift the bottleneck to balancing computation and acceptance length. RecGuide uses completed verification history to select each request's next block size $B_t\in\{8,16,32\}$.
Let $e_t$ average verifier entropy over causally valid positions, excluding those after the first mismatch. We combine two-round certainty with the previous block outcome $z_t$:

\begin{equation}
s_t=-\frac{e_{t-1}+e_{t-2}}{2}, \qquad
z_t=(B_{t-1},F_{t-1}),
\label{eq:recguide-state}
\end{equation}

where $F_{t-1}$ indicates whether the previous block was fully accepted. 
While $s_t$ mainly reflects whether a long block should shrink under high uncertainty, $z_t$ further encourages a larger drafting budget after full acceptance, since the true acceptance potential may extend beyond the current block boundary. 
Given this, RecGuide estimates the expected accepted progress $V_B(s_t,z_t)$ from calibration data for each candidate $B$. Following the hardware-aware cost modeling of DSpark, an offline profile provides the execution cost $T_C(B)$ under serving load $C$. RecGuide then selects

\begin{equation}
B_t^*
=
\arg\max_{B\in \{8,16,32\}}
\left[
V_B(s_t,z_t)-\rho_C T_C(B)
\right],
\label{eq:recguide-routing}
\end{equation}

where $\rho_C$ weights the profiled execution cost at load $C$. A larger block is chosen only when its expected acceptance gain justifies the additional computation. 
The decision is resolved through lightweight table lookup at runtime, introducing negligible overhead.

\subsection{System Implementation}
\label{sec:implementation}

Algorithmic optimization alone does not directly translate into end-to-end gains. For draft-verify overlap, naively batching the verifier and prospective branches causes redundant prefix KV reads and padding due to unequal sequence lengths. We therefore extend SGLang with shared-prefix KV access and padding-free execution, allowing all branches to reuse the canonical prefix while processing only valid query tokens. We also redesign the mapping of mixed attention workloads onto GPU SMs.

Dynamic block sizing faces a similar issue: mixing different block sizes within a batch forces shorter requests to pad to the largest block, offsetting their computational savings. RecGuide therefore adopts block-size bucketed serving. After each verification round, requests are grouped by their planned block size, and the scheduler forms batches only from requests in the same bucket, preserving the efficiency of adaptive block sizing while keeping request states and canonical KV independent.

%% file: docs/6-experiments.tex
\input{docs/6-experiments-tables}

\section{Experiments}
\label{sec:experiments}

\subsection{Experimental Setup}
\label{sec:setup}

\textbf{Models and Benchmarks.} 
We evaluate RecGuide on Nemotron-Labs-Diffusion (NLD) 8B and 14B \citep{fu2026nemotronlabsdiffusion}, a representative state-of-the-art diffusion--AR self-speculation family. Fixing the backbone enables fair decoding-paradigm comparisons with LinearSS and its quadratic variant TiDAR \citep{liu2026tidar}, isolating runtime gains from architectural differences.
Our evaluation covers eight benchmarks spanning three categories: \textbf{Math} (GSM8K \citep{cobbe2021trainingverifierssolvemath}, MATH-500 \citep{lightman2024let}, AIME25 \citep{aime25}); \textbf{Code} (HumanEval \citep{chen2021evaluatinglargelanguagemodels}, MBPP \citep{austin2021programsynthesislargelanguage}, LiveCodeBench \citep{jain2025livecodebench}); \textbf{QA \& instruction following} (GPQA \citep{rein2023gpqagraduatelevelgoogleproofqa}, IFEval \citep{zhou2023instructionfollowingevaluationlargelanguage}). We compare against vanilla LinearSS, DFlash, and TiDAR. Due to training budget constraints, DFlash is trained on ShareGPT \citep{sharegpt52k} following the SpecForge recipe \citep{li2026specforgeflexibleefficientopensource}, while other configurations follow official settings.

\textbf{Evaluation Settings and Metrics.} All experiments are conducted on a single NVIDIA B200 GPU with SGLang \citep{zheng2024sglang} engine and FlashInfer \citep{ye2025flashinfer} backend. We evaluate our two RecGuide strategies under their corresponding serving regimes. Following NLD's recipe, all methods use greedy decoding. We set concurrency $C=1$ and block size $B=16$ for low-concurrency experiments. Potential mismatch positions are identified by $M_i<\delta$, with $\delta=0.5$ throughout all experiments. For higher-concurrency serving, we vary concurrency over $C\in \{2,4,8,16,32,64,128\}$ and compare RecGuide against fixed block sizes $B\in\{8,16,32\}$ using both aggregate and per-request throughput.

\subsection{Low-Concurrency Performance}
\label{sec:low-concurrency performance}

\ExperimentMainResultsTable

\ExperimentBranchReuseTable

Table~\ref{tab:main-results} compares RecGuide with representative diffusion-AR speculative decoding paradigms under the single-request setting, reporting average acceptance length $\tau$ and end-to-end speedup over autoregressive decoding. Although LinearSS accepts substantially more tokens per round than DFlash, its two full-backbone forwards limit end-to-end speedup, allowing DFlash with a lightweight drafter to remain faster. RecGuide, by contrast, achieves the highest speedup, averaging $4.06\times$ on NLD-8B and $4.40\times$ on NLD-14B over autoregressive decoding, and delivering up to $1.8\times$ speedup over vanilla LinearSS on AIME25. Under the same two-forward accounting, RecGuide raises the effective acceptance length to more than 14 tokens, approaching the $B=16$ limit of serial execution.

Longer acceptance, however, does not necessarily imply faster decoding even after removing the two-forward dependency of linear self-speculation, as TiDAR illustrates. Its quadratic speculative layout prepares a future draft for every possible acceptance position, incurring $O(B^2)$ speculative computation. Consequently, its longer acceptance does not translate into higher end-to-end speedup because of the increased forward cost, leaving TiDAR behind LinearSS on average. Moreover, TiDAR drafts from hypothesized acceptance positions without explicitly supplying a correction token at likely mismatches. The repeated draft may still be incorrect; even after a successful correction, later tokens in the same parallel draft cannot condition on it, potentially degrading the draft quality.

RecGuide instead assigns the second-ranked candidate to selected risky positions, providing an explicit corrected prefix once verified. Despite using far fewer prospective branches than TiDAR, RecGuide retains high mismatch coverage, as shown in Table~\ref{tab:overlap-branch-reuse}. Correction and continuation branches cover most decoding rounds, leaving only 24.85\% and 27.22\% of rounds on NLD-8B and NLD-14B, respectively, unmatched by any branch. Furthermore, among matched correction branches, the second-ranked candidate matches the verifier correction in over $60\%$ of cases, supporting explicit correction as a reliable drafting signal. Continuation branches are even more reusable: after full-block acceptance, the first draft token almost always matches the additional next token produced by the verifier. RecGuide therefore preserves acceptance lengths comparable to TiDAR with far less speculative computation, ultimately achieving the highest decoding efficiency.

\subsection{High-Concurrency Performance}
\label{sec:high-concurrency performance}

\begin{figure}[t]
    \centering

    \begin{subfigure}[b]{0.45\linewidth}
        \centering
        \includegraphics[
            width=\linewidth
        ]{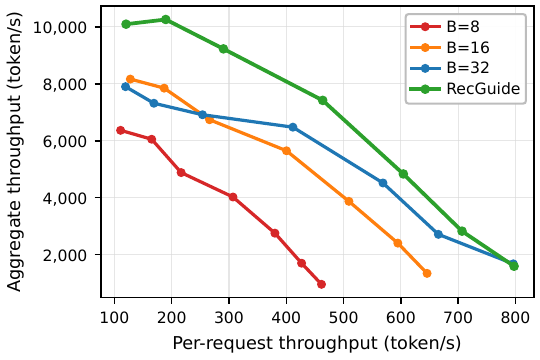}
        \caption{}
        \label{fig:throughput-8b}
    \end{subfigure}
    \hfill
    \begin{subfigure}[b]{0.45\linewidth}
        \centering
        \includegraphics[
            width=\linewidth
        ]{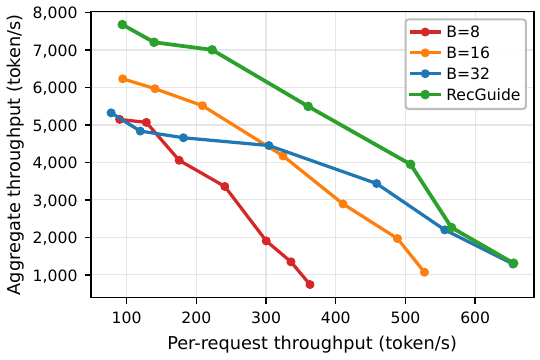}
        \caption{}
        \label{fig:throughput-14b}
    \end{subfigure}

    \caption{
    Aggregate and per-request throughput trade-offs under different concurrency levels on (a) NLD-8B and (b) NLD-14B. We compare fixed block sizes $B\in\{8,16,32\}$ with RecGuide.
    }
    \label{fig:throughput-tradeoff}
\end{figure}

\ExperimentBlockSelectionTable

Figure~\ref{fig:throughput-tradeoff} evaluates RecGuide under higher serving loads, showing the trade-off between aggregate and per-request throughput across $C\in \{2,4,8,16,32,64,128\}$. Increasing concurrency raises aggregate throughput at the expense of per-request throughput. Compared with fixed $B\in \{8,16,32 \}$, RecGuide favors larger blocks when compute remains available and shifts toward smaller blocks as execution becomes compute limited. This adaptive allocation consistently improves aggregate throughput on both NLD-8B and NLD-14B, advancing the overall serving frontier.

Table~\ref{tab:dynamic-block-selection} makes this load adaptation explicit. On NLD-8B, the fraction of $B=32$ decreases from 88.13\% at $C=2$ to 9.75\% at $C=128$, while $B=8$ increases from 9.33\% to 78.21\%; NLD-14B exhibits the same transition. This confirms that RecGuide adjusts drafting budgets with serving load rather than relying on a single fixed configuration. Block-size distributions also vary across task categories at the same concurrency. In particular, QA \& IF shifts toward smaller blocks earlier than MATH and CODE, consistent with higher verification uncertainty and demonstrating request-level adaptation beyond load awareness alone.

These distributions also explain the per-request throughput trends in Figure~\ref{fig:throughput-tradeoff}. At $C=2$ and $C=128$, RecGuide predominantly selects $B=32$ and $B=8$, respectively, so its per-request throughput approaches the corresponding fixed-block baselines, whereas the more diverse assignments at intermediate concurrency provide greater gains from request-specific adaptation. Notably, fixed $B=8$ remains inferior to $B=16$ in aggregate throughput at the highest concurrency, since shorter blocks require more decoding rounds and amplify progress imbalance across concurrent requests. RecGuide avoids this drawback by selecting short blocks only when beneficial and regrouping requests by their planned block size at each round, allowing aggressive use of $B=8$ without inheriting the inefficiency of a globally fixed short-block policy.

\subsection{Ablation Study}
\label{sec:ablation}

We further decompose RecGuide by progressively adding the key components of Predictive Draft-Verify Overlap and Verification Guided Adaptive Drafting to vanilla LinearSS under their respective load regimes. We report aggregate throughput in tokens per second (TPS) on NLD-8B / 14B.

\ExperimentAblationTableone

\textbf{Predictive Draft-Verify Overlap.} Table~\ref{tab:draft_verify_overlap_ablation} ablates the prospective branches at $C=1$. Starting with the first correction branch, we add the second correction and continuation branches to isolate the benefits of broader mismatch coverage and full-block continuation. The first correction branch alone improves throughput by 34.29\% and 28.72\% on NLD-8B / 14B, respectively. The second correction branch yields a smaller gain, which we attribute to later risky positions being less causally grounded and harder to predict, supporting our decision to bound the number of correction branches. Adding the continuation branch further raises the improvement to 51.76\% and 49.36\%. These results show that three prospective branches capture most of the overlap benefit while keeping computation bounded.

\ExperimentAblationTabletwo

\textbf{Verification Guided Adaptive Drafting.} We next ablate the adaptive block policy over $C\in\{2,4,8,16,32,64,128\}$, as shown in Table~\ref{tab:dynamic_block_ablation}. Fixed Block follows NLD's official recipe, using a concurrency-specific block size shared by all requests. Adding two-round entropy $s_t$ improves TPS by 10.49\% / 7.60\% on NLD-8B / 14B. The previous block size and full-acceptance state $z_t$ further guide budget expansion. Request-specific block sizes introduce unequal draft lengths within batches, so padding to the largest block can offset these computational savings. Block-size bucketing groups requests by their selected size, translating compute savings into serving throughput gains. Together, adaptive selection and bucketed execution improve TPS by 39.84\% and 32.80\%, respectively.

%% file: docs/6-experiments-tables.tex
\newlength{\tauentrywidth}
\newlength{\speedentrywidth}
\newlength{\entrywidth}

\newcommand{\tauentry}[1]{%
\makebox[\entrywidth][l]{#1}%
}
\newcommand{\speedentry}[1]{%
\makebox[\entrywidth][l]{#1$\times$}%
}

\newcommand{\spdmetric}{\cellcolor{gray!25}Speedup}
\newcommand{\spdcell}[1]{\cellcolor{gray!25}\speedentry{#1}}

\newcommand{\hdr}[1]{\multirow{2}{*}{\raisebox{-0.6ex}{#1}}}

\newcommand{\gain}[1]{%
{\scriptsize\textcolor{green!50!black}{(+#1\%)}}%
}

\newcommand{\ExperimentMainResultsTable}{%
\begin{table}[t]
\centering
\caption{
Average acceptance length ($\tau$) per two-forward round and decoding speedup over autoregressive decoding on NLD-8B/14B at $C=1$, $B=16$. RecGuide can verify two blocks in this window.
}
\label{tab:main-results}

\scriptsize
\setlength{\tabcolsep}{3.2pt}
\renewcommand{\arraystretch}{1.0}

\settowidth{\tauentrywidth}{18.19}
\settowidth{\speedentrywidth}{5.49$\times$}
\setlength{\entrywidth}{\tauentrywidth}
\ifdim\speedentrywidth>\entrywidth
    \setlength{\entrywidth}{\speedentrywidth}
\fi

\resizebox{\linewidth}{!}{%
\begin{tabular}{c l c c c c c c c c c c}
\toprule
\hdr{Model}
& \hdr{Method}
& \hdr{Metric}
& \multicolumn{3}{c}{\textsc{Math}}
& \multicolumn{3}{c}{\textsc{Code}}
& \multicolumn{2}{c}{\textsc{QA \& IF}}
& \textsc{Overall} \\

    \cmidrule(lr){4-6}
    \cmidrule(lr){7-9}
    \cmidrule(lr){10-11}
    \cmidrule(lr){12-12}

    & &
    & GSM8K & MATH-500 & AIME25
    & HumanEval & MBPP & LCB
    & GPQA & IFEval
    & \textit{Avg.} \\

    \midrule

    \multirow{8}{*}{NLD-8B}

    & \multirow{2}{*}{LinearSS}
    & $\tau$
    & \tauentry{9.94}
    & \tauentry{11.21}
    & \tauentry{10.49}
    & \tauentry{9.71}
    & \tauentry{8.15}
    & \tauentry{9.78}
    & \tauentry{10.62}
    & \tauentry{9.57}
    & \tauentry{9.93} \\

    & & \spdmetric
    & \spdcell{3.03}
    & \spdcell{2.95}
    & \spdcell{2.42}
    & \spdcell{2.91}
    & \spdcell{2.67}
    & \spdcell{2.36}
    & \spdcell{2.58}
    & \spdcell{2.53}
    & \spdcell{2.68} \\

    & \multirow{2}{*}{DFlash}
    & $\tau$
    & \tauentry{4.26}
    & \tauentry{4.31}
    & \tauentry{4.90}
    & \tauentry{4.28}
    & \tauentry{3.78}
    & \tauentry{4.64}
    & \tauentry{5.11}
    & \tauentry{3.90}
    & \tauentry{4.40} \\

    & & \spdmetric
    & \spdcell{2.80}
    & \spdcell{2.87}
    & \spdcell{3.13}
    & \spdcell{2.83}
    & \spdcell{2.47}
    & \spdcell{3.09}
    & \spdcell{3.39}
    & \spdcell{2.57}
    & \spdcell{2.89} \\

    & \multirow{2}{*}{TiDAR}
    & $\tau$
    & \tauentry{13.44}
    & \tauentry{15.80}
    & \tauentry{14.54}
    & \tauentry{12.92}
    & \tauentry{9.18}
    & \tauentry{13.48}
    & \tauentry{15.18}
    & \tauentry{12.52}
    & \tauentry{13.38} \\

    & & \spdmetric
    & \spdcell{2.62}
    & \spdcell{2.87}
    & \spdcell{2.34}
    & \spdcell{2.52}
    & \spdcell{1.88}
    & \spdcell{2.32}
    & \spdcell{2.57}
    & \spdcell{2.26}
    & \spdcell{2.42} \\

    & \multirow{2}{*}{RecGuide}
    & $\tau$
    & \tauentry{13.95}
    & \tauentry{16.08}
    & \tauentry{15.68}
    & \tauentry{13.64}
    & \tauentry{11.08}
    & \tauentry{13.54}
    & \tauentry{14.84}
    & \tauentry{13.33}
    & \tauentry{14.02} \\

    & & \spdmetric
    & \spdcell{4.05}
    & \spdcell{4.66}
    & \spdcell{4.38}
    & \spdcell{4.07}
    & \spdcell{3.22}
    & \spdcell{3.91}
    & \spdcell{4.27}
    & \spdcell{3.94}
    & \spdcell{4.06} \\

    \midrule

    \multirow{8}{*}{NLD-14B}

    & \multirow{2}{*}{LinearSS}
    & $\tau$
    & \tauentry{9.74}
    & \tauentry{10.50}
    & \tauentry{9.35}
    & \tauentry{11.99}
    & \tauentry{10.81}
    & \tauentry{9.27}
    & \tauentry{10.79}
    & \tauentry{8.35}
    & \tauentry{10.10} \\

    & & \spdmetric
    & \spdcell{2.84}
    & \spdcell{3.06}
    & \spdcell{2.36}
    & \spdcell{4.15}
    & \spdcell{3.41}
    & \spdcell{2.50}
    & \spdcell{2.82}
    & \spdcell{2.43}
    & \spdcell{2.95} \\

    & \multirow{2}{*}{DFlash}
    & $\tau$
    & \tauentry{4.78}
    & \tauentry{4.37}
    & \tauentry{3.81}
    & \tauentry{4.93}
    & \tauentry{5.49}
    & \tauentry{4.14}
    & \tauentry{5.28}
    & \tauentry{3.80}
    & \tauentry{4.58} \\

    & & \spdmetric
    & \spdcell{3.13}
    & \spdcell{2.94}
    & \spdcell{2.50}
    & \spdcell{3.30}
    & \spdcell{3.43}
    & \spdcell{2.80}
    & \spdcell{3.54}
    & \spdcell{2.57}
    & \spdcell{3.03} \\

    & \multirow{2}{*}{TiDAR}
    & $\tau$
    & \tauentry{13.10}
    & \tauentry{16.24}
    & \tauentry{13.90}
    & \tauentry{17.72}
    & \tauentry{15.80}
    & \tauentry{13.26}
    & \tauentry{17.24}
    & \tauentry{12.20}
    & \tauentry{14.94} \\

    & & \spdmetric
    & \spdcell{2.74}
    & \spdcell{3.09}
    & \spdcell{2.47}
    & \spdcell{3.78}
    & \spdcell{3.13}
    & \spdcell{2.48}
    & \spdcell{3.08}
    & \spdcell{2.36}
    & \spdcell{2.89} \\

    & \multirow{2}{*}{RecGuide}
    & $\tau$
    & \tauentry{13.61}
    & \tauentry{15.43}
    & \tauentry{13.41}
    & \tauentry{18.19}
    & \tauentry{12.56}
    & \tauentry{13.27}
    & \tauentry{16.21}
    & \tauentry{10.97}
    & \tauentry{14.21} \\

    & & \spdmetric
    & \spdcell{4.22}
    & \spdcell{4.82}
    & \spdcell{4.21}
    & \spdcell{5.49}
    & \spdcell{3.74}
    & \spdcell{4.17}
    & \spdcell{5.03}
    & \spdcell{3.55}
    & \spdcell{4.40} \\

    \bottomrule
\end{tabular}%
}

\end{table}
}

\newcommand{\ExperimentBranchReuseTable}{%
\begin{table}[t]
\centering
\caption{
Branch outcomes in draft-verify overlap. Correction and Continuation report coverage / reuse rate (\%); No Hit denotes rounds unmatched by any prospective branch.
}
\label{tab:overlap-branch-reuse}

\small
\setlength{\tabcolsep}{5.0pt}
\renewcommand{\arraystretch}{1.05}

\begin{tabular}{c c c c c c}
\toprule

Model
& Branch Outcome
& \textsc{Math}
& \textsc{Code}
& \textsc{QA \& IF}
& \textsc{Overall} \\

\midrule

\multirow{3}{*}{NLD-8B}

& Correction
& 32.46 / 21.46
& 41.63 / 25.81
& 33.31 / 17.25
& 36.11 / 22.04 \\

& Continuation
& 43.85 / 42.92
& 27.89 / 27.14
& 48.55 / 47.88
& 39.04 / 38.24 \\

& No Hit
& 23.68
& 30.48
& 18.14
& 24.85 \\

\midrule

\multirow{3}{*}{NLD-14B}

& Correction
& 32.37 / 21.98
& 30.99 / 20.38
& 35.44 / 18.61
& 32.62 / 20.54 \\

& Continuation
& 37.12 / 35.95
& 43.81 / 36.53
& 39.22 / 38.88
& 40.15 / 36.90 \\

& No Hit
& 30.51
& 25.20
& 25.33
& 27.22 \\

\bottomrule
\end{tabular}

\end{table}
}

\newcommand{\ExperimentBlockSelectionTable}{%
\begin{table}[t]
\centering
\caption{
Dynamic block-size distributions across serving loads. Each entry reports the percentage of requests selecting $B=8/16/32$, respectively; Overall is the macro average over eight benchmarks.
}
\label{tab:dynamic-block-selection}

\fontsize{8pt}{9.0pt}\selectfont
\setlength{\tabcolsep}{5pt}
\renewcommand{\arraystretch}{1.0}

\begin{tabular}{c c c c c c}
\toprule

Model
& Concurrency
& \textsc{Math}
& \textsc{Code}
& \textsc{QA \& IF}
& \textsc{Overall} \\

\midrule

\multirow{4}{*}{NLD-8B}
& 2
& 3.61 / 1.28 / 95.11
& 4.63 / 1.51 / 93.86
& 24.97 / 5.97 / 69.07
& 9.33 / 2.54 / 88.13 \\

& 8
& 0.63 / 26.91 / 72.46
& 0.66 / 39.18 / 60.16
& 12.85 / 48.73 / 38.43
& 3.70 / 36.96 / 59.34 \\

& 32
& 28.83 / 52.95 / 18.22
& 40.45 / 47.70 / 11.86
& 60.76 / 22.35 / 16.89
& 41.17 / 43.33 / 15.50 \\

& 128
& 72.46 / 17.25 / 10.30
& 82.03 / 11.11 / 6.87
& 81.12 / 5.65 / 13.23
& 78.21 / 12.04 / 9.75 \\

\midrule

\multirow{4}{*}{NLD-14B}
& 2
& 1.40 / 3.11 / 95.49
& 0.32 / 1.92 / 97.76
& 12.14 / 12.83 / 75.03
& 3.68 / 5.09 / 91.23 \\

& 8
& 0.56 / 42.26 / 57.18
& 0.04 / 37.14 / 62.82
& 6.27 / 58.77 / 34.97
& 1.79 / 44.47 / 53.74 \\

& 32
& 28.31 / 63.75 / 7.94
& 20.26 / 69.51 / 10.23
& 56.77 / 31.81 / 11.43
& 32.40 / 57.92 / 9.67 \\

& 128
& 84.07 / 11.56 / 4.37
& 83.80 / 10.21 / 5.99
& 83.41 / 5.94 / 10.66
& 83.81 / 9.65 / 6.55 \\

\bottomrule
\end{tabular}

\end{table}
}

\newcommand{\ExperimentAblationTableone}{%
\begin{wraptable}{r}{0.46\linewidth}
\centering
\vspace{-0.8\baselineskip}
\caption{
Ablation of prospective branches in Predictive Draft-Verify Overlap at $C=1$.
TPS is averaged over all benchmarks.
}
\label{tab:draft_verify_overlap_ablation}

\small
\setlength{\tabcolsep}{5.0pt}
\renewcommand{\arraystretch}{1.08}

\begin{tabular}{l c c}
\toprule
Method
& NLD-8B
& NLD-14B \\
\midrule

LinearSS
& 595
& 470 \\

+ 1st Correction
& 799 \gain{34.29}
& 605 \gain{28.72} \\

+ 2nd Correction
& 811 \gain{36.30}
& 612 \gain{30.21} \\

+ Continuation
& 903 \gain{51.76}
& 702 \gain{49.36} \\

\bottomrule
\end{tabular}

\end{wraptable}
}

\newcommand{\ExperimentAblationTabletwo}{%
\begin{wraptable}{r}{0.46\linewidth}
\centering
\vspace{-0.8\baselineskip}
\caption{
Ablation of verification guided adaptive drafting. TPS is averaged over all benchmarks and concurrency levels.
}
\label{tab:dynamic_block_ablation}

\small
\setlength{\tabcolsep}{4.5pt}
\renewcommand{\arraystretch}{1.08}

\begin{tabular}{l c c}
\toprule
Method
& NLD-8B
& NLD-14B \\
\midrule

Fixed Block
& 4725
& 3756 \\

+ $s_t$
& 5221 \gain{10.49}
& 4041 \gain{7.60} \\

+ $z_t$
& 5254 \gain{11.19}
& 4071 \gain{8.40} \\

+ Bucketing
& 6608 \gain{39.84}
& 4987 \gain{32.80} \\

\bottomrule
\end{tabular}

\end{wraptable}
}